\documentclass[final]{cvpr}

\usepackage{color}
\usepackage{ifthen}
\usepackage{float}
\usepackage{stfloats}
\usepackage{alltt}
\usepackage{newlfont} % for Box
\usepackage{wrapfig}
\usepackage{booktabs}
\usepackage{multirow}
\usepackage{comment}
\usepackage{gensymb}

\usepackage{balance}
\usepackage{comment}
\usepackage{amsmath}
\usepackage{amssymb}
\usepackage{amsthm}
\usepackage{times}
\usepackage{epsfig}
\usepackage{graphicx}
\usepackage{amsmath}
\usepackage{amssymb}
\usepackage{amsthm}
\usepackage{subcaption}
\usepackage{microtype}
\usepackage{blindtext}
\usepackage{array}

\usepackage[title]{appendix}
\usepackage{chngcntr}
\usepackage{flushend}
\usepackage{balance}

\newcommand{\revision}[1]{{\color{black} #1}}

\definecolor{DeltaColor}{rgb}{0.039,0.73,0.71}
\definecolor{SetaColor}{rgb}{0.867, 0.0235, 0.376}
\definecolor{SigmaColor}{rgb}{0.98,0.45,0.0}
\definecolor{AlphaColor}{rgb}{0,0,0.8}
\definecolor{BetaColor}{rgb}{0.8,0,0.8}
\definecolor{GammaColor}{rgb}{0.5,0,0.7}
\definecolor{EpsilonColor}{rgb}{0.353,0.725,0.906}
\definecolor{TauColor}{rgb}{0.423,0.235,0.192}

\newcommand{\nothing}[1]{}

\definecolor{AudioColor}{rgb}{0.56,0.34,0.62}

\definecolor{DeadlineColor}{rgb}{0.9,0.4,0} % energetic color

\definecolor{figred}{rgb}{1,0,0}
\definecolor{figgreen}{rgb}{0,0.6,0}
\definecolor{figblue}{rgb}{0,0,1}
\definecolor{figpink}{rgb}{1,0.63,0.63}

\newcounter{pccount}
\floatstyle{plain}
\newcommand{\filename}[1]{\url{#1}}
\newcommand{\foldername}[1]{\url{#1}}

\DeclareMathOperator*{\argmin}{argmin}         % for argmin or argmax equations

\usepackage[pagebackref=true,breaklinks=true,colorlinks,bookmarks=false]{hyperref}

\newcommand*{\affaddr}[1]{#1} 
\newcommand*{\affmark}[1][*]{\textsuperscript{#1}}
\newcommand*{\email}[1]{\small{\texttt{#1}}}

\newcommand\blfootnote[1]{%
  \begingroup
  \renewcommand\thefootnote{}\footnote{#1}%
  \addtocounter{footnote}{-1}%
  \endgroup
}

\def\cvprPaperID{1058}
\def\confYear{CVPR 2021}

\graphicspath{
{figures}
{images}
}

\begin{document}
\title{SCANimate: Weakly Supervised Learning of Skinned Clothed Avatar Networks}
\author{Shunsuke Saito\affmark[1]\textsuperscript{*} \hspace{0.3in} 
Jinlong Yang\affmark[1] \hspace{0.3in} 
Qianli Ma\affmark[1,2]\hspace{0.3in}
Michael J. Black\affmark[1]
\\
\affaddr{\affmark[1]Max Planck Institute for Intelligent Systems, T\"ubingen, Germany} \quad \affaddr{\affmark[2]ETH Z\"urich}\\
% Max Planck Institute for Intelligent Systems, T\"ubingen, Germany
\email{\{ssaito,jyang,qma,black\}@tuebingen.mpg.de}
}

\twocolumn[{%
\renewcommand\twocolumn[1][]{#1}%
\maketitle
\begin{center}
    \newcommand{\teaserwidth}{\textwidth}
\vspace{-0.2in}
    \centerline{
    \includegraphics[width=1\linewidth]{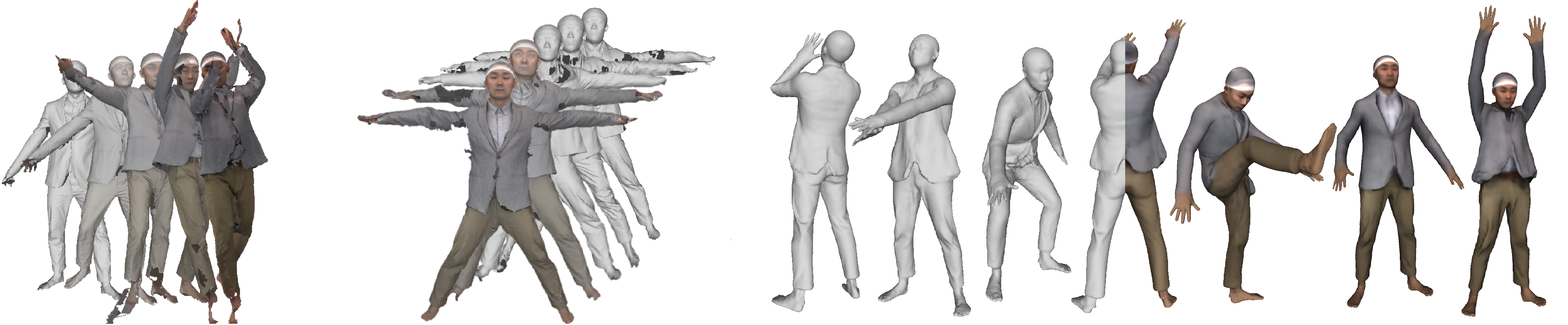}
     }
   \vspace{-5pt}
   \captionof{figure}{\textbf{SCANimate.} Given a set of raw scans with multiple poses containing self-intersections, holes, and noise (left), SCANimate automatically aligns all scans to a canonical pose (middle) and learns a Scanimat, a fully animatable avatar that produces pose-dependent deformations and texture without garment-specific templates or mesh registration (right).}
\label{fig:teaser}
\end{center}%
}]

%%%%%%%%% ABSTRACT
\begin{abstract}{~}
We present SCANimate, an end-to-end trainable framework that takes raw 3D scans of a clothed human and turns them into an animatable avatar. These avatars are driven by pose parameters and have realistic clothing that moves and deforms naturally. SCANimate does not rely on a customized mesh template or surface mesh registration. We observe that fitting a parametric 3D body model, like SMPL, to a clothed human scan is tractable while surface registration of the body topology to the scan is often not, because clothing can deviate significantly from the body shape.
\blfootnote{$^*$Currently at Facebook Reality Labs.}
We also observe that articulated transformations are invertible, resulting in geometric cycle-consistency in the posed and unposed shapes. These observations lead us to a weakly supervised learning method that aligns scans into a canonical pose by disentangling articulated deformations without template-based surface registration. Furthermore, to complete missing regions in the aligned scans while modeling pose-dependent deformations, we introduce a locally pose-aware implicit function that learns to complete and model geometry with learned pose correctives. In contrast to commonly used global pose embeddings, our local pose conditioning significantly reduces long-range spurious correlations and improves generalization to unseen poses, especially when training data is limited. Our method can be applied to pose-aware appearance modeling to generate a fully textured avatar. We demonstrate our approach on various clothing types with different amounts of training data, outperforming existing solutions and other variants in terms of fidelity and generality in every setting.
The code is available at \small{\url{https://scanimate.is.tue.mpg.de}}.
\end{abstract}

\section{Introduction}
\label{sec:intro}

%Goal
Parametric models of 3D human bodies are widely used for the analysis and synthesis of human shape, pose, and motion.
While existing models typically represent ``minimally clothed" bodies \cite{anguelov2005scape, hasler2009statistical, loper2015smpl, osman2020star, xu2020ghum}, many applications require realistically clothed bodies.
Our goal is to make it easy to produce a  realistic 3D avatar of a clothed person that can be reposed and animated as easily as existing models like SMPL \cite{loper2015smpl}.
In particular, the model must support clothing that moves and deforms naturally, with detailed 3D wrinkles, and the rendering of realistically textured images.

To that end, we introduce \textit{SCANimate (Skinned Clothed Avatar Networks for animation)}, which creates high-quality animatable clothed humans,  called Scanimats, from raw 3D scans. 
SCANimate has the following properties: (1) we learn an articulated clothed human model directly from raw scans, completely eliminating the need for surface registration of a custom template or synthetic clothing simulation data, (2) our parametric model retains the complex and detailed deformations of clothing present in the original scans such as wrinkles and sliding effects of garments with arbitrary topology, (3) a Scanimat can be animated directly using SMPL pose parameters, and (4) our approach predicts pose-dependent clothing deformations based on local pose parameters, providing generalization to unseen poses. 

Recent data-driven approaches have shown promise for learning parametric models of clothed humans from real-world observations \cite{lahner2018deepwrinkles, CAPE:CVPR:20, patel20tailornet,pons2017clothcap}.
However, these approaches typically limit the supported clothing types and topology because
they require accurate surface registration of a common  template mesh to 3D training scans \cite{lahner2018deepwrinkles, CAPE:CVPR:20, pons2017clothcap}.
Concurrent work by Ma et al.~\cite{SCALE:CVPR:21} learns clothing deformation without surface registration, yet it is unclear if the method works on raw scans with noise and holes.
Learning from real-world observations is essentially challenging because raw 3D scans are un-ordered point clouds with missing data, changing topology, multiple clothing layers, and  sliding motions between the body and garments.
Although one can learn from synthetic data generated by physics-based clothing simulation \cite{guan2012drape,gundogdu2019garnet, patel20tailornet}, the results are less realistic, the data preparation is time consuming and non-trivial to scale to the real-world clothing. 

To address these issues, SCANimate learns directly from raw scans of people in clothing.
Body scanning is becoming common, and scans can be obtained from a variety of devices.
Scans contain high-frequency details, capture varied clothing topology, and are inherently realistic.
To make learning from scans possible, we make several contributions: canonicalization, implicit skinning fields, cycle consistency, and implicit shape learning.

\vspace{5pt}
\noindent
{\bf Canonicalization and Implicit Skinning Fields.} 
The first step involves transforming the raw scans to a common pose so we can learn to model pose-dependent surface deformations (e.g.~bulging, stretching, wrinkling, and sliding), i.e.~pose ``correctives".
But we are not seeking a traditional ``registration" of the scans to a common mesh topology, since this is, in general, not feasible with clothed bodies.
Instead, we learn continuous functions of 3D space that allow us to transform posed scans to a canonical pose and back again.

The key idea is to build this on linear blend skinning (LBS), which
traditionally defines weights on the surface of a mesh that encode how much each vertex is influenced by the rotation of a body joint.
To deal with raw scans of unknown topology, we extend this notion by defining skinning weights {\em implicitly} everywhere in 3D space. 
Specifically, given a 3D location $\boldsymbol{x}$, we regress a continuous vector function $g$ represented by a neural network, $g(\boldsymbol{x}):\mathbb{R}^{3} \to \mathbb{R}^{J}$, which defines the skinning weights.
An inverse LBS function uses the regressed skinning weights to ``undo" the pose of the body and transforms the points into the canonical space. 
As this representation makes no assumptions about the topology or resolution of input scans, we can \textit{canonicalize} arbitrary non-watertight meshes.
Furthermore, we can easily generate animations of the parametric clothed avatar by applying forward LBS to the clothed body in the canonical pose with the learned pose correctives.

\noindent
{\bf Cycle Consistency.}
Despite the desirable properties of canonicalization, learning the skinning function is ill-posed since we do not have ground truth training data that specifies the weights.
To address this, we exploit two key observations.
First, as demonstrated in previous work \cite{hu2020learning,yang2016estimation,Zhang_2017_CVPR}, fitting a parametric human body model such as SMPL \cite{loper2015smpl} to 3D scans is more tractable than surface registration.
We leverage SMPL's skinning weights, which are defined only on the body surface, to regularize our more general skinning function.
Second, the transformations between the posed space and the canonical space should be cycle-consistent. 
Namely, inverse LBS and forward LBS together should form an identity mapping as illustrated in Fig.~\ref{fig:cycle}, which provides a self-supervision signal for training the skinning function.
After training the skinning function, we obtain the canonicalized scans (all in the same pose).

\noindent
{\bf  Learning Implicit Pose Correctives.}
Given the canonicalized scans, we learn a model that captures the pose-dependent deformations.
However a problem remains: the original raw scans often contain holes, and so do the canonicalized scans.
To deal with this and with the arbitrary topology of clothing, we use an implicit surface representation \cite{chen2018implicit_decoder,mescheder2019occupancy, park2019deepsdf}.
As multiple canonicalized scans will miss different regions, with this approach, they complement each other, while retaining details present in the original inputs. 
Furthermore, unlike traditional approaches \cite{lahner2018deepwrinkles,CAPE:CVPR:20, patel20tailornet,yang2018analyzing}, where pose-dependent deformations are conditioned on entire pose parameters, we spatially filter out irrelevant pose features from the input conditions by leveraging the learned skinning weights. 
In this way, we effectively prune long-range spurious correlations between garment deformations and body joints, achieving plausible pose correctives for unseen poses even from a small number of training scans.
The resulting learned Scanimat can be easily reposed and animated with SMPL pose parameters.

In summary, our main contributions are (1) the first end-to-end trainable framework to build a high-quality parametric clothed human model from raw scans, (2) a novel weakly-supervised formulation with geometric cycle-consistency that disentangles articulated deformations from the local pose correctives without requiring ground-truth training data, and (3) a locally pose-aware implicit surface representation that models pose-dependent clothing deformation and generalizes to unseen poses. 
Our results show that SCANimate is superior to existing solutions in terms of generality and accuracy.
Furthermore, we perform an extensive study to evaluate the technical contributions that are critical for success.
\revision{
The code and example Scanimats can be found at \small{\url{https://scanimate.is.tue.mpg.de}}.
}

\section{Related Work}
\label{sec:related_work}

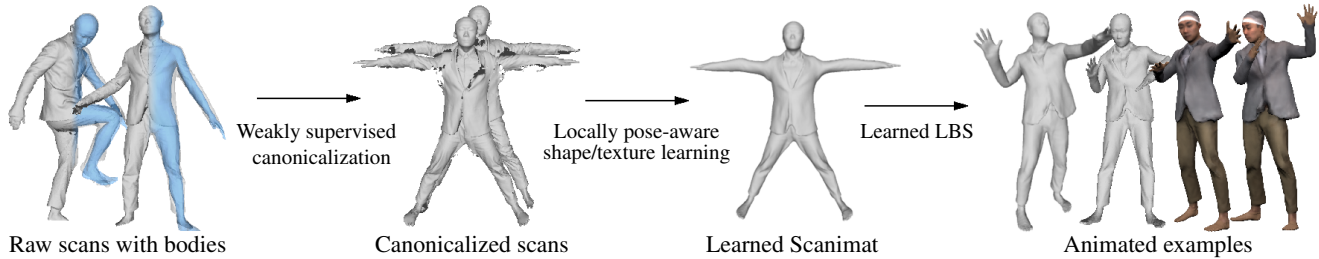
\begin{figure*}[ht!]
\centering
\includegraphics[width=\linewidth]{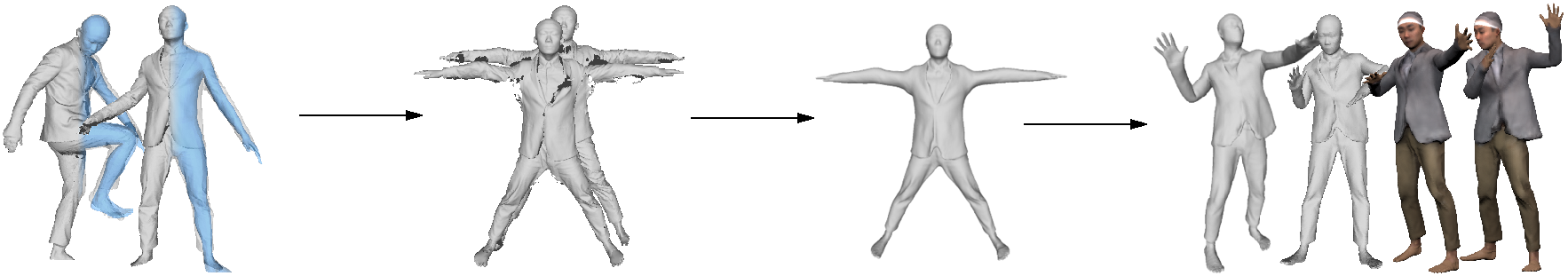}
\put(-410,36){\footnotesize{Weakly supervised}}
\put(-404,26){\footnotesize{canonicalization}}
\put(-290,36){\footnotesize{Locally pose-aware}}
\put(-294,28){\footnotesize{shape/texture learning}}
\put(-174,36){\footnotesize{Learned LBS}}
\put(-496,-7){\small{Raw scans with bodies \qquad  \qquad \qquad Canonicalized scans \qquad \qquad \quad \; Learned Scanimat \qquad \qquad \qquad \quad \; Animated examples}}
\vspace{-5pt}
\caption{\small{
\textbf{Overview.} SCANimate learns a pose-aware parametric clothed human model directly from raw scans in a weakly supervised manner. The resulting Scanimats can be animated with SMPL pose parameters, producing realistic pose-dependent deformations and texture.}}
\label{fig:overview}
\end{figure*}

\noindent
\textbf{Parametric Models for Human Bodies and Clothing.}
Parametric body models \cite{anguelov2005scape, hasler2009statistical, loper2015smpl, osman2020star, xu2020ghum} learn statistical body shape variations and pose-dependent shape correctives that capture non-linear body deformation and compensate for linear blend skinning artifacts \cite{allen2002articulated, james2005skinning, kry2002eigenskin, kurihara2004modeling,lewis2000pose}. While these approaches achieve high-fidelity and intuitive control of human body shape and pose, they only focus on bodies without clothing. Similar ideas have been extended to model clothed bodies by introducing additional garment layers \cite{Danerek17DeepGarment, de2010stable, guan2012drape, gundogdu2019garnet, Kim:2013:NEP, lahner2018deepwrinkles, xu2014sensitivity} or adding displacements or transformations to the base human body mesh \cite{alldieck19cvpr, alldieck2018video, CAPE:CVPR:20, Neophytou2014layered, yang2018analyzing}. These parametric clothed human models decompose garment deformations into articulated deformations and local deformations such that pose correctives only focus on non-rigid local deformations. Thus, it is essential to obtain the inverse skinning transformation \cite{patel20tailornet} by using the surface registration of a well-defined template \cite{lahner2018deepwrinkles, loper2015smpl, CAPE:CVPR:20, yang2018analyzing, Zhang_2017_CVPR} or using synthetic simulation data \cite{SMPLicit:2021, guan2012drape, gundogdu2019garnet, patel20tailornet}. 
However, these requirements limit the applicability of the approaches to fairly simple clothing, with a fixed topology, and without complex interactions between garments and the body.

In contrast, our work uses a weakly supervised approach to build a parametric clothed human model from raw scans without the requirement of a template and surface registration. We canonicalize posed scans and learn an implicit surface with arbitrary topology \cite{gropp2020implicit} conditioned on pose parameters by leveraging a fitted human body model to the scan data \cite{bualan2008naked, bhatnagar2020combining,yang2016estimation,yu2018doublefusion,Zhang_2017_CVPR}. Moon et al.~\cite{moon2020deephandmesh} similarly propose a weakly supervised method for learning a fine-grained hand model from scan data by deforming a fitted base hand model \cite{romero2017embodied}; the approach is non-trivial to extend to human clothing with varying topology. 

The most related work to ours is Neural Articulated Shape Approximation (NASA) \cite{deng2019neural}, where the composition of occupancy networks \cite{chen2018implicit_decoder,mescheder2019occupancy} articulated by the fitted SMPL model are directly learned from posed scans in the same spirit as structured implicit functions \cite{genova2020local,genova2019learning}. \revision{Concurrent work, LEAP~\cite{LEAP:CVPR:21}, extends  a similar framework to a multi-subject setting.}
Through an extensive study in Sec.~\ref{sec:evaluation}, we find that the compositional implicit functions proposed in \cite{deng2019neural} are more prone to artifacts and less generalizable to unseen poses than our LBS-based formulation. 

\noindent
\textbf{Pose Canonicalization via Inverse LBS.}
The key to successful canonicalization is learning transformations in the form of skinning weights in a continuous space. Learning skinning weights for varied topologies has become possible using neural networks with graph convolutions \cite{bertiche2020deepsd,liu2019neuroskinning,RigNet}. Given a neutral-posed template, these networks predict skinning weights together with a skeleton \cite{RigNet} or pose-dependent deformations \cite{bertiche2020deepsd}. While they predict skinning weights on a neutral-posed template in a fully supervised manner, our problem requires learning skinning weights, not only on the surface mesh, but in both the canonical and posed space without ground-truth skinning weights. 

Extending LBS skinning weights from an underlying body model to the continuous space is used in the data preparation step of ARCH \cite{huang2020arch} and LoopReg \cite{bhatnagar2020loopreg}. However, in these approaches, the skinning weights are uniquely determined by the underlining body and not \textit{learnable}. We argue, and experimentally demonstrate, in Sec.~\ref{sec:evaluation} that jointly learning skinning weights leads to visually pleasing canonicalization while maximizing the reproducibility of input scans by the reconstructed parametric model. Inspired by recent unsupervised methods using cycle consistency \cite{chang2018pairedcyclegan,zhu2017unpaired}, we leverage geometric cycle consistency between the canonical space and posed space to learn skinning weights in a weakly supervised manner without requiring any ground-truth training data. Concurrent work, FTP~\cite{PTF:CVPR:21}, proposes a similar idea but is limited to body modeling; instead, we extend the traditional LBS to the entire 3D space and enable clothing surface modeling.

\noindent
\textbf{Reconstructing Clothed Humans.}
Reconstructing humans from depth maps \cite{chibane2020implicit, wang2020normalgan,yu2018doublefusion}, images \cite{bogo2016keep, kanazawa2018end, kolotouros2019learning}, or video \cite{kanazawa2019learning, kocabas2020vibe} is also extensively studied. While many works focus on the minimally clothed human body \cite{bogo2016keep, guan2009estimating, kanazawa2018end, kolotouros2019learning, lassner2017unite}, recent approaches show promise in reconstructing clothed human models from RGB inputs using the SMPL mesh with displacements \cite{alldieck19cvpr,alldieck2018video, zhu2019detailed}, external garment layers \cite{bhatnagar2019mgn, jiang2020bcnet}, depth maps \cite{gabeur2910moulding, smith2019facsimile}, voxels \cite{varol2018bodynet, Zheng2019DeepHuman}, or implicit functions \cite{huang2020arch, saito2019pifu, saito2020pifuhd}. However, these approaches do not learn, or infer, pose-dependent deformation of garments, and simply apply articulated deformations to the reconstructed shapes.
This results in unrealistic pose-dependent deformations that lack garment specific wrinkles. 
Our work differs by focusing on learning pose-dependent clothing deformation from scans. 

\section{Method}
\label{sec:method}
Figure \ref{fig:overview} shows an overview of our pipeline. 
The input is a set of raw 3D scans of a person in clothing, together with fitted minimally clothed body models.
Here we use the SMPL model \cite{loper2015smpl} fit to the scans to obtain body joints and blend skinning weights, which we exploit in learning.
Given the input, we first learn bidirectional transformations between the posed space and canonical space by predicting skinning weights as a function of space coordinates (Sec.~\ref{sec:lbs}).
To address the lack of ground truth correspondence of the scan data, we leverage geometric cycle consistency to learn continuous skinning functions. 
The raw scans are canonicalized with the learned bidirectional transformations.
We further learn a locally pose-aware signed distance function, parameterized by a neural network, from canonicalized scans using implicit geometric regularization \cite{gropp2020implicit} (Sec.~\ref{sec:igr}). 
For implementation details, including hyper parameters and network architectures, see Appendix \ref{sup:sec:implementation_details}.  

\setlength{\abovedisplayskip}{5pt}
\setlength{\belowdisplayskip}{5pt}
\subsection{Canonicalization}
\label{sec:lbs}

\begin{figure}[t]
\centering
\includegraphics[width=\linewidth]{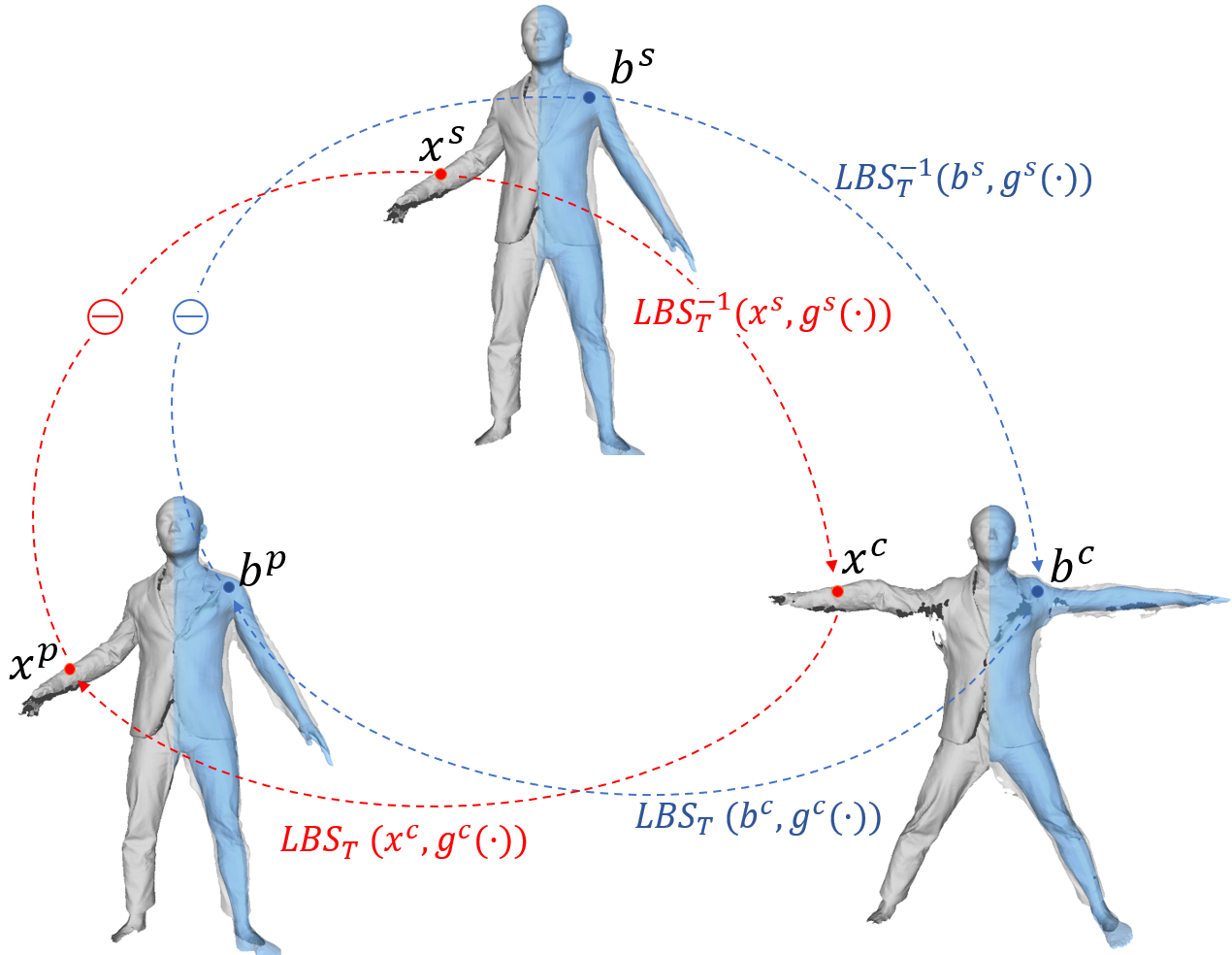}
\put(-150, 88){\small{Input scan}}
\put(-72, -8){\small{Canonicalized}}
\put(-218, -9){\small{Reposed}}
\vspace{-5pt}
\caption{\small{
\textbf{Canonicalization with cycle consistency.} The geometric cycle consistency loss, with the guidance from the underlining body model, leads to successful canonicalization.}}
\label{fig:cycle}
\end{figure}
Instead of a traditional skinning scheme that assigns a skinning weight vector $\boldsymbol{w} \in \mathbb{R}^J$, where $J$ is the number of joints, to each point on a surface, we extend the notion of skinning using a continuous function: we train a model that takes any point in the space as input and outputs its skinning weight vector $\boldsymbol{w}$. Figure \ref{fig:cycle} illustrates the principles. 
We specifically focus on points from two surfaces, the clothing surface $\mathbf{X}$ and the  body surface  $\mathbf{B}$. 
To be more specific about the canonicalization step, let us first define the posed space and the canonical space. The posed space is defined for each scan, and the canonical space is shared across all the scans. Let $\mathbf{X}^s_i=\{\boldsymbol{x}^s \in \mathbb{R}^3\}$ be vertices on the original scan in the posed space, where $i$ is the frame index of the scans, and $\mathbf{X}_i^c=\{\boldsymbol{x}^c \in \mathbb{R}^3\}$ be vertices on the unposed scans in the canonical space, which are not known. Now we seek a mapping function that aligns the posed scans in a canonical pose. While the mapping function can be arbitrarily defined, we observe that this can be formulated as a composition of the \textit{known} rigid transformations of body joints, $\mathbf{T}^i=\{\mathbf{T}^i_j \in SE(3), j=1, \ldots, J\}$, which come from the fitted SMPL model. More specifically, given a set of blending weights $\boldsymbol{w}$, we define linear blend skinning ($LBS$) and inverse linear blend skinning ($LBS^{-1}$) functions as follows:
\begin{equation}
\begin{gathered}
\label{eq:lbs}
   \boldsymbol{X}^p_i = LBS_{\mathbf{T}_i}(\boldsymbol{X}^c_i, \boldsymbol{w}(\boldsymbol{X}^c_i))=(\sum{w_j \mathbf{T}_{i,j}}) \boldsymbol{X}^c_i\\
   \boldsymbol{X}^c_i = LBS^{-1}_{\mathbf{T}_i}(\boldsymbol{X}^s_i, \boldsymbol{w}(\boldsymbol{X}_i^s)) = (\sum{w_j \mathbf{T}_{i,j}})^{-1}  \boldsymbol{X}_i^s,
\end{gathered}
\end{equation}
where $\mathbf{X}^p_i=\{\boldsymbol{x}^p \in \mathbb{R}^3\}$ are the vertices of the reposed scans and ideally should have the same value as $\mathbf{X}^s_i$. The LBS function maps arbitrary points in the canonical space to the posed space represented by $\mathbf{T}_i$ and the inverse LBS function maps points in the posed space to the canonical space. In other words, the equations above show that given skinning weights $\boldsymbol{w}$ on vertices, we can not only apply any pose to the canonicalized shapes as in a traditional character animation pipeline \cite{lewis2000pose}, but also transform back the posed shapes into the canonical space.

\noindent\textbf{Implicit Skinning Fields.}
In contrast to traditional applications, where the skinning weights for each point are predefined, either by artists or by automatic methods \cite{baran2007automatic,huang2020arch, yang2018analyzing}, skinning weights on the raw scan data are not known {\em a priori}. Fortunately, we can learn them in a weakly supervised manner, such that all the scans can be decomposed into articulated deformations and non-rigid deformations.

To this end, we introduce two neural networks called the \textit{forward skinning net} and the \textit{inverse skinning net}:
\begin{equation}
\begin{gathered}
\label{eq:skin_net}
    \boldsymbol{w}(\boldsymbol{x}^c_i) = g^c_{\Theta_1}(\boldsymbol{x}^c_i):\mathbb{R}^{3} \to \mathbb{R}^{J}\\
    \boldsymbol{w}(\boldsymbol{x}^s_i) = g^s_{\Theta_2}(\boldsymbol{x}^s_i, \boldsymbol{z}^s_i):\mathbb{R}^{3} \times \mathbb{R}^{\mathcal{Z}_s} \to \mathbb{R}^{J},
\end{gathered}
\end{equation}
where $\boldsymbol{z}^s_i$ represents a latent embedding, and $\Theta_1$ and $\Theta_2$ are the learnable parameters of the multilayer perceptrons (MLP), which we omit below for notational brevity. The forward skinning net predicts LBS skinning weights of queried 3D locations in the canonical space. Similarly, the inverse skinning net predicts skinning weights in the posed space of each training scan. 
Notably, this continuous representation is advantageous over other alternatives including fully connected networks and graph convolutional networks \cite{liu2019neuroskinning, CAPE:CVPR:20, ranjan2018generatingcoma} as it does not depend on a fixed number of vertices or predefined topology. \revision{Empirically we observe that jointly learning $\boldsymbol{z}^s_i$ in an auto-decoding fashion \cite{park2019deepsdf} leads to superior performance compared to taking pose parameters as input;} \revision{see Appendix~\ref{sup:sec:discussion} for discussion.}

By combining Eq.~\ref{eq:lbs} and \ref{eq:skin_net}, we can compute the mappings between the canonical and posed spaces via: 
\begin{equation}
\begin{gathered}
\label{eq:lbs2}
    \boldsymbol{x}^p_i = LBS_{\mathbf{T}_i}(\boldsymbol{x}^c_i, g^c(\boldsymbol{x}^c_i))\\
    \boldsymbol{x}^c_i = LBS^{-1}_{\mathbf{T}_i}(\boldsymbol{x}^s_i, g^s(\boldsymbol{x}^s_i, \boldsymbol{z}^s_i)).
\end{gathered}
\end{equation}
Note that these functions are differentiable.

\noindent\textbf{Learning Skinning.}
To successfully train $g^c(\cdot)$ and $g^s(\cdot)$ without ground truth weights on the scans, we leverage two key observations: (1) the regions close to the human body model are highly correlated with the nearest body parts where ground-truth skinning weights are available; (2) any points in the posed space should be mapped back to the same points after reapplying LBS to the canonicalized points. To utilize (1), we use the underlying SMPL body model's LBS skinning weights as guidance for the canonical and posed space. More specifically, $g^s(\cdot)$ and $g^c(\cdot)$ at points on the scans are loosely guided by the nearest neighbor point on the body model and its SMPL skinning weights, propagating skinning weights from body models to the input scans. 

Most importantly, observation (2) plays a central role in the success of the weakly supervised learning. It allows us to formulate cycle consistency constraints, updating both $g^c(\cdot)$ and $g^s(\cdot)$ such that wrongly associated skinning weights that break the cycle consistency are highly penalized. Our evaluation in Sec.~\ref{sec:evaluation} shows that the cycle consistency constraints are critical to decompose articulated deformations. Note that the jointly learned $g^c(\cdot)$ is used to learn and animate the pose-aware clothed human model (see Sec.~\ref{sec:igr}).

Our final objective function is defined as:
\begin{equation}
\label{eq:cano}
\begin{split}
    E_{cano}&(\Theta_1, \Theta_2, \left\{ \boldsymbol{z}^s_i \right\})=\\
    &\sum_{i}{(\lambda_{B}E_{B}+\lambda_{S}E_{S}+E_{C}+E_{R})},
\end{split}
\end{equation}
where $E_{B}$ and $E_{S}$ are body-guided loss functions, $E_{C}$ is based on cycle consistency, and $E_{R}$ is a regularization term. $E_{B}$ ensures $g^c(\cdot)$ and $g^p(\cdot)$ predict SMPL skinning weights on the body surface by
\begin{equation}
\label{eq:body_guided}
\begin{split}
E_{B} = & \sum_{ \boldsymbol{b}^c_i \in \mathbf{B}^c_i }{ \|g^c(\boldsymbol{b}^c_i)-\boldsymbol{w'}(\boldsymbol{b}^c_i)\| } \\
&+ \sum_{ \boldsymbol{b}^s_i \in \mathbf{B}^s_i }{ \|g^s(\boldsymbol{b}^s_i,\boldsymbol{z}^s_i)-\boldsymbol{w'}(\boldsymbol{b}^s_i)\| },
\end{split}
\end{equation}
where $\mathbf{B}^c_i=\{\boldsymbol{b}^c_i\in \mathbb{R}^3\}$ and $\mathbf{B}^s_i=\{\boldsymbol{b}^s_i\in \mathbb{R}^3\}$ are vertices on the canonical and posed body surfaces, and $\boldsymbol{w'}(\cdot)$ are the SMPL LBS weights. 
\revision{See Appendix~\ref{sup:sec:implementation_details} for details to obtain $\mathbf{B}^c_i$.}
Similarly, $E_{S}$ is the regression loss between the predicted weights and the LBS weights on the nearest neighbor body vertex:
\begin{multline}
E_{S}=\sum_{\boldsymbol{x}^s_i \in \mathbf{X}^s_i}(\|g^s(\boldsymbol{x}^s_i,\boldsymbol{z}^s_i)-\boldsymbol{w'}(\argmin_{\boldsymbol{b}^s_i\in \mathbf{B}^s_i}{d(\boldsymbol{x}^s_i,\boldsymbol{b}^s_i)})\| \\
+ \|g^c(\boldsymbol{x}^c_i)-\boldsymbol{w'}(\argmin_{\boldsymbol{b}^s_i\in \mathbf{B}^s_i}{d(\boldsymbol{x}^s_i,\boldsymbol{b}_i^s)})\|).
\end{multline}
Note that this nearest neighbor assignment is also used in \cite{huang2020arch} for training data preparation. However, in Sec.~\ref{sec:evaluation}, we show that this alone is prone to inaccurate assignments, causing severe artifacts.

We facilitate cycle consistency with two terms. $E_{C'}$ directly constrains the consistency of skinning weights between the canonical space and the posed space, and $E_{C''}$ facilitates cycle consistency on the vertices of the posed meshes as follows:
\begin{gather}
    E_{C} = \lambda_{C'} E_{C'} + \lambda_{C''} E_{C''}\\
    E_{C'}=\sum_{\boldsymbol{x}^s_i \in \mathbf{X}^s_i}{\|g^s(\boldsymbol{x}^s_i,\boldsymbol{z}^s_i)-g^c(\boldsymbol{x}^c_i)\|}\\
    E_{C''}=\sum_{\boldsymbol{x}^s_i \in \mathbf{X}^s_i}{\|\boldsymbol{x}^p_i- \boldsymbol{x}^s_i\|}.
\end{gather}
Notice that cycle consistency can hold only if we start from the posed space since points in the canonical space can be mapped to the same location in case of self-intersection. 

Lastly, our regularization term consists of a sparsity constraint $E_{Sp}$, a smoothness term $E_{Sm}$, and a statistical regularization on the latent code $E_{Z}$ as follows:
\begin{gather}
    E_{R}=\lambda_{Sp} E_{Sp}+\lambda_{Sm} E_{Sm} + \lambda_{Z} E_{Z},\\
    E_{Sp}=\sum_{\boldsymbol{x}^s_i}{|g^s(\boldsymbol{x}^s_i, \boldsymbol{z}^s_i)|^\beta} \quad \beta=0.8,\\
    E_{Sm}=\sum_{\boldsymbol{e} \in \mathbf{E} / \mathbf{C}}{\|g^s(\boldsymbol{e}_1,\boldsymbol{z}^s_i) - g^s(\boldsymbol{e}_2,\boldsymbol{z}^s_i)\|},\\
    E_{Z}=\|\boldsymbol{z}^p_i\|^2_2,
\end{gather}
where $\boldsymbol{e}=(\boldsymbol{e}_1, \boldsymbol{e}_2)$, $\mathbf{E}$ is the set of edges on the triangulated scans and we mask out concave regions $\mathbf{C}$ so that skinning weights are not propagated across merged body parts due to self-intersection (\revision{See Appendix~\ref{sup:sec:implementation_details} for details.}). 

After training, we canonicalize all the scans by applying the inverse LBS transform (Eq.~\ref{eq:lbs2}) to all vertices on the scans. By eliminating triangles with large distortion (see Appendix~\ref{sup:sec:implementation_details} for details), we obtain the canonical scans used to learn a pose-aware parametric clothed human model. 
\subsection{Locally Pose-aware Implicit Shape Learning}
\label{sec:igr}
Given the canonicalized partial scans together with the learned skinning weights, we learn a parametric clothed human model with pose-aware deformations. To this end, we base our shape representation on an implicit surface representation \cite{chen2018implicit_decoder, mescheder2019occupancy, park2019deepsdf} as it supports arbitrary topology with fine details. However, real scans have holes and such partial scans cause difficulty obtaining ground truth occupancy labels since the meshes are not water-tight. To handle partial scans as input, we learn a signed distance function $f_\Phi(\boldsymbol{x})$ based on a multilayer perceptron (for brevity, we omit the network parameters $\Phi$), using implicit geometric regularization (IGR) \cite{gropp2020implicit} by minimizing the following objective function:
\begin{gather}
    E_{shape}(\Phi)=\sum_i{(E_{LS}+\lambda_{igr} E_{IGR} + \lambda_o E_{O})}\\ 
    E_{LS} = \sum_{\boldsymbol{x} \in \mathbf{X'}^c_i}\left(\left|f\left(\boldsymbol{x}\right)\right|+\left\|\nabla_{\boldsymbol{x}} f\left(\boldsymbol{x}\right)-\boldsymbol{n}(\boldsymbol{x})\right\|\right),\\
    E_{IGR} = \mathbb{E}_{\boldsymbol{x}}\left(\left\| \nabla_{\boldsymbol{x}} f(\boldsymbol{x}) \right\|-1\right)^{2},
\end{gather}
\begin{gather}
    E_{O} = \mathbb{E}_{\boldsymbol{x}}\left(\exp(-\alpha \cdot \left|f\left(\boldsymbol{x}\right)\right|)\right) \quad \alpha \gg 1,
\end{gather}
where $E_{LS}$ ensures the zero level-set of the predicted SDF lies on the given points with its surface normal aligned with that of the input scans, $\boldsymbol{n}(\boldsymbol{x})$. $E_{IGR}$ is the Eikonal regularization term that regularizes the function $f$ to satisfy the Eikonal equation $\left\|\nabla_{\boldsymbol{x}} f(\cdot)\right\|=1$. $E_O$ regularizes off-surface SDF values from being close to the level-set surface as in \cite{sitzmann2019siren}. Remarkably, this formulation does not require ground truth signed distance for non-surface points and naturally fills in the missing regions by leveraging the inductive bias of multilayer perceptrons as shown in \cite{gropp2020implicit}. 

To learn pose-dependent deformations of clothing, we could condition the function $f$ with the pose features $\theta \in \mathbb{R}^{J \times 4}$ (we use quaternions as in \cite{osman2020star}). 
However, the  straightforward approach of concatenating the pose features with Cartesian coordinates, namely $f(\boldsymbol{x}, \theta)$,  suffers from  overfitting due to the limited pose variations in the training data and spurious correlations between joints. 
Since the relationship between body joints and clothing deformation tends to be non-local \cite{xu2014sensitivity}, it is also important to limit the influence of irrelevant joints to reduce spurious correlations \cite{osman2020star}. 
Thus, we need an attention mechanism to associate spatial locations with only the relevant pose features. To this end, we modify the function $f$:
\begin{equation}
\label{eq:localpose}
    f(\boldsymbol{x}, (W \cdot g^c(\boldsymbol{x})) \circ \theta), W \in \mathbb{R}^{J\times J},
\end{equation}
where $g^c(\cdot)$ is the skinning network learned in Sec.~\ref{sec:lbs}, $W$ is the weight map that converts skinning weights into pose attention weights\revision{, and $\circ$ denotes element-wise product}. Specifically, if we want a 3D point that is skinned to the $n^{th}$ joint with non-zero skinning weights to pay attention to the $m^{th}$ joint, $W_{m,n}$ and $W_{n,m}$ are set to 1, otherwise, they are set to 0. The weight map is essential because the movement of one joint will be propagated to regions associated with neighboring body joints (e.g.~raising the shoulders lifts up an entire T-shirt). In this paper, we set $W_{n,m} = 1$ when $n^{th}$ joint is within $4$-ring neighbors of $m^{th}$ joint in the kinematic tree. By reducing spurious correlations, our formulation significantly reduces over-fitting artifacts given a set of unseen poses,  demonstrating better generalization ability even with a small number of input scans (see Sec.~\ref{sec:evaluation}).

\section{Experimental Results}
\label{sec:result}

\paragraph{Dataset and Metric.}
For evaluation and comparison with baseline methods, we use the CAPE dataset~\cite{CAPE:CVPR:20}, which includes raw 3D scan sequences and SMPL model fits.

We evaluate generalization to unseen poses with both pose interpolation (denoted as Int. in tables) and extrapolation tasks (denoted as Ex. in tables). The motion sequences are randomly split into training (80\%) and test (20\%) sets, where the test sequences are used to evaluate extrapolation. For the training sequences, we choose every 10th frame starting from the first frame as training scans and every 10th frame with 5 frame strides from the training sequences for the interpolation evaluation. 
We perform Marching Cubes~\cite{lorensen1987marching} to the predicted implicit surface in canonical space as in Eq~\ref{eq:localpose} and then pose it by forward LBS in Eq.~\ref{eq:lbs} to get the resulting meshes.
For quantitative evaluation, we use scan-to-mesh distance $D_{s2m}$ (cm) and surface normal consistency $D_n$, where a nearest neighbor vertex on the resulting meshes is used to compute the average L2 norm.

In addition, we conduct a perceptual study to assess the plausibility score, $P$, of generated garment shapes and deformations. Workers on Amazon Mechanical Turk (AMT) are given a pair of side-by-side images or videos showing a rendered result from our approach and another approach; the left-right order of the results is randomized.
The task is to choose the result with the most realistic clothing.
We continue this $N$ times and compute the probability of the other approach being favored $P=M/N$, where $M$ is how many times the users chose the other method over ours. 
In other words, we set our approach as baseline with a constant score $P = 0.5$; for other approaches, if $P < 0.5$, ours achieves higher fidelity. 
The perceptual score for image and video pairs is denoted as $P_i$ and $P_v$,  respectively. 
While $P_i$ focuses on the plausibility of static clothing, $P_v$ reveals the temporal consistency and realism of pose-dependent clothing deformations. Note that we provide only the perceptual scores for the extrapolation task as numerical evaluation is difficult due to the stochasticity of clothing deformations. 

\subsection{Evaluation}
\label{sec:evaluation}

\noindent
{\bf   Canonicalization.}
The goal of canonicalization is to disentangle articulated deformations from other non-rigid deformations for effective shape learning. We choose two baseline approaches to replace our canonicalization module. The first, as used by \cite{huang2020arch}, copies skinning weights on the clothed scans from the nearest neighbor body vertex. 
The other approach is based on weighted correspondences by interpolating skinning weights from the k-nearest neighbors (we use $k=6$) in the spirit of \cite{yang2018analyzing}.
This reduces the impact of a wrong clothing-body association that limits the performance of single nearest neighbor assignment.
% 0.012518054658869297 0.3008801213877174
% 0.012450229971611444 0.29874306142355384

% \begin{table}[h]
%     \centering
%     \begin{tabular}{|c|c|c|c|c|}
%     \hline
%         \multicolumn{2}{|c|}{}& Ours & NN~\cite{huang2020arch} & KNN\\
%         \hline
%         \multirow{4}{*}{Int.} & $D_{s2m}$ $\downarrow$ & \textbf{5.70} & 12.5 & 12.5 \\
%         \cline{2-5}
%         & $D_n$ $\downarrow$ & \textbf{0.253} & 0.301 & 0.299 \\
%         \cline{2-5}
%         & $P_i$ $\uparrow$ & \textbf{1.00} & 0.598 & 0.657\\
%         \cline{2-5}
%         & $P_v$ $\uparrow$ & \textbf{1.00} & 0.770 & 0.758\\
%         \hline
%         \multirow{2}{*}{Ex.} & $P_i$ $\uparrow$ & \textbf{1.00} & 0.355 & 0.454\\
%         \cline{2-5}
%         & $P_v$ $\uparrow$ & \textbf{1.00} & 0.646 & 0.816\\
%         \hline
%     \end{tabular}
%     \caption{Comparison of canonicalization with KNN based approaches.}
%     \label{tab:canonicalization}
% \end{table}

\begin{table}[t!]
    \centering
    \caption{Quantitative comparison of canonicalization. \revision{As the perceptual score is pair-wise and compared against ours, we put 0.5 for the proposed approach throughout the tables. $D_{s2m}$ is in centimeters throughout the tables.}}
        \vspace{-0.1in} {\small
    \begin{tabular}{lp{1.5cm}p{1.2cm}<{\centering}p{1.2cm}<{\centering}p{1.2cm}<{\centering}}
    \toprule
        \multicolumn{2}{c}{}& Ours & NN~\cite{huang2020arch} & KNN\\
        \midrule
        \multirow{4}{*}{Int.} & $D_{s2m}$ $\downarrow$ & \textbf{0.570} & 1.25 & 1.25 \\
        % \cline{2-5}
        & $D_n$ $\downarrow$ & \textbf{0.253} & 0.301 & 0.299 \\
        % \cline{2-5}
        & $P_i$ $\uparrow$ & \textbf{0.5} & 0.374 & 0.396\\
        % \cline{2-5}
        & $P_v$ $\uparrow$ & \textbf{0.5} & 0.435 & 0.431\\
        \midrule
        \multirow{2}{*}{Ex.} & $P_i$ $\uparrow$ & \textbf{0.5} & 0.262 & 0.312\\
        % \cline{2-5}
        & $P_v$ $\uparrow$ & \textbf{0.5} & 0.392 & 0.449\\
    \bottomrule
    \end{tabular}}
    \label{tab:canonicalization}
%    \vspace{-5pt}
\end{table} 
\begin{figure}[t]
\centering
\includegraphics[width=\linewidth]{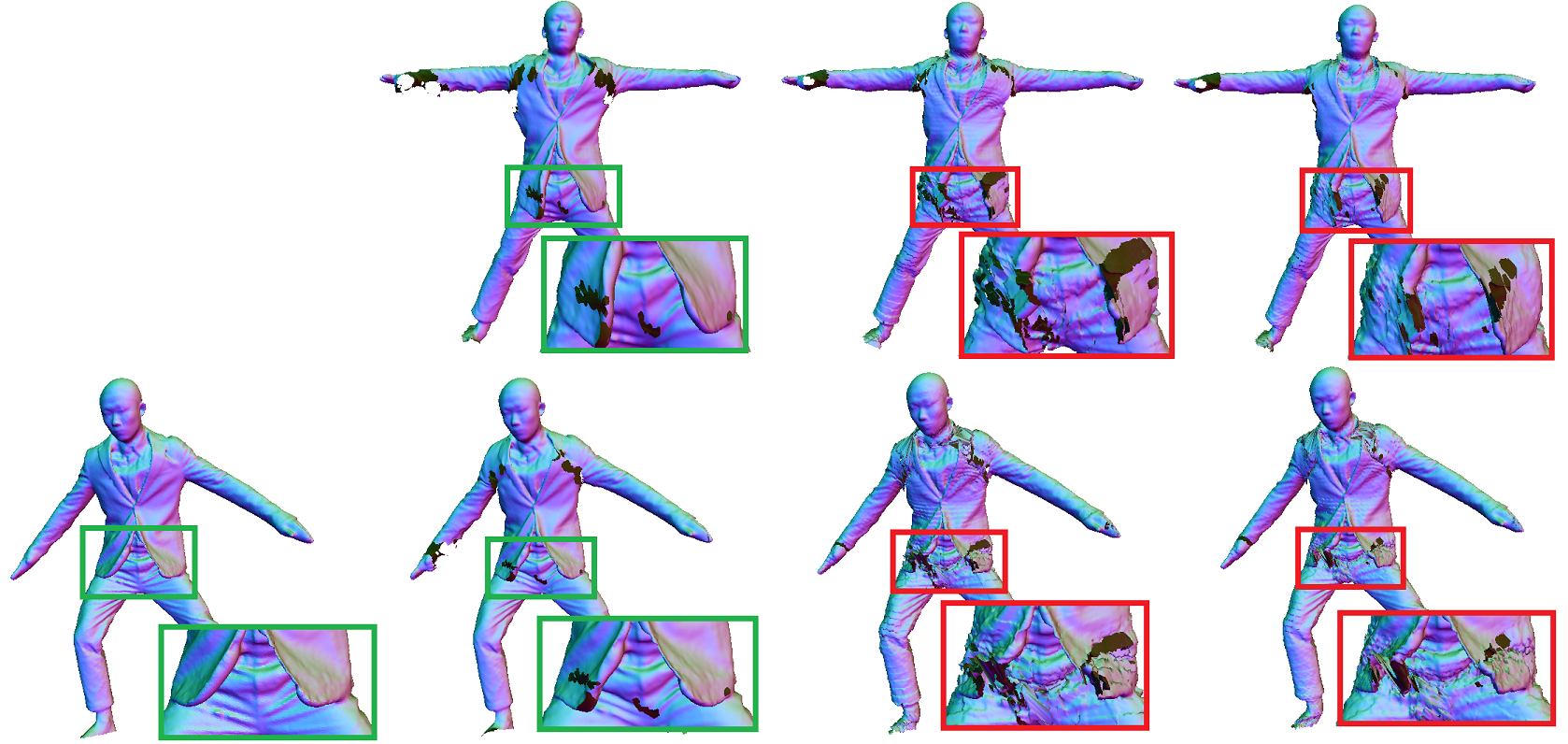}
\put(-227,-8){\small{Ground truth \qquad \; Ours \qquad \qquad NN \qquad \qquad \; KNN }}
\vspace{-5pt}
\caption{\small{
\textbf{Qualitative comparison on canonicalization.} Top: canonicalization results. Bottom: reposed canonicalization results. Compared with our method, the baseline methods suffer from severe artifacts.}}
\label{fig:eval_canonicalization}
\end{figure}

Figure \ref{fig:eval_canonicalization} shows that the two baseline methods break the cycle consistency with wrong associations of the skinning weights, resulting in noticeable artifacts.
The inaccurate canonicalization results are propagated to the parametric model learning, substantially degrading the quality of reconstructed avatars as shown in Tab.~\ref{tab:canonicalization}. Our approach with cycle consistency successfully normalizes the input scans into a canonical pose while retaining coherent geometric details, enabling the parametric modeling of clothed avatars. 

\noindent
{\bf Locally Pose-aware Shape Learning.}
We evaluate our local pose representation using the learned skinning weights for pose-dependent shape learning and compare against commonly used global pose conditioning~\cite{deng2019neural,lahner2018deepwrinkles,CAPE:CVPR:20,patel20tailornet,yang2018analyzing}. To this end, we replace the second input of Eq.~\ref{eq:localpose} with the global pose parameter, $\theta$, as a baseline. To assess the generalization ability, both models are trained on 100\%, 50\%, 10\% and 5\% of the original training set.

\begin{table}[t]
    \centering
    \caption{Quantitative evaluation of the importance of locality in the pose conditioning on different sizes of training data.}
        \vspace{-0.1in}{\small 
    \begin{tabular}{p{0.4cm}p{1.2cm}p{1.0cm}<{\centering}p{1.0cm}<{\centering}p{1.0cm}<{\centering}p{1.0cm}<{\centering}}
    \toprule
        \multicolumn{2}{c}{Train size (\%)}& 100 & 50 & 10 & 5 \\
        \hline
        \multicolumn{6}{c}{Local pose conditioning (Ours)}\\
        \hline
        \multirow{4}{*}{Int.} & $D_{s2m}$ $\downarrow$ & \textbf{0.570}  & 0.663 & 0.699 & 0.732\\
        % \cline{2-6}
        & $D_n$  $\downarrow$ & \textbf{0.253} & 0.253 & 0.261 & 0.268\\
        % \cline{2-6}
        & $P_i$   $\uparrow$ & \textbf{0.5} & 0.476 & 0.466 & 0.398\\
        % \cline{2-6}
        & $P_v$   $\uparrow$ & \textbf{0.5} & 0.453 & 0.435 &0.425\\
        \hline
        \multirow{2}{*}{Ex.} & $P_i$ $\uparrow$ & \textbf{0.5} & 0.429 & 0.359 & 0.359\\
        % \cline{2-6}
        & $P_v$  $\uparrow$ & \textbf{0.5} & 0.408 & 0.408 & 0.343\\
        \hline
        \multicolumn{6}{c}{Global pose conditioning}\\
        \hline
        \multirow{3}{*}{Int.} & $D_{s2m}$ $\downarrow$ & 0.768 & 0.786 & 1.54 & 2.38\\
        % \cline{2-6}
        & $D_n$  $\downarrow$ & 0.253 & 0.256 & 0.293 & 0.354\\
        % \cline{2-6}
        & $P_i$   $\uparrow$ & 0.424& 0.393 & 0.350 & 0.252\\
        % \cline{2-6}
        & $P_v$   $\uparrow$ & 0.468 & 0.457 & 0.363 & 0.301\\
        \hline
        \multirow{2}{*}{Ex.} & $P_i$ $\uparrow$ & 0.417  & 0.401 & 0.291 & 0.192\\
        % \cline{2-6}
        & $P_v$  $\uparrow$ & 0.436 & 0.382 & 0.311 & 0.203\\
    \bottomrule
    \end{tabular}}
    \label{tab:eavl_poseCondition}
 %       \vspace{-5pt}
\end{table}
\begin{figure}[t]
\centering
\includegraphics[width=\linewidth]{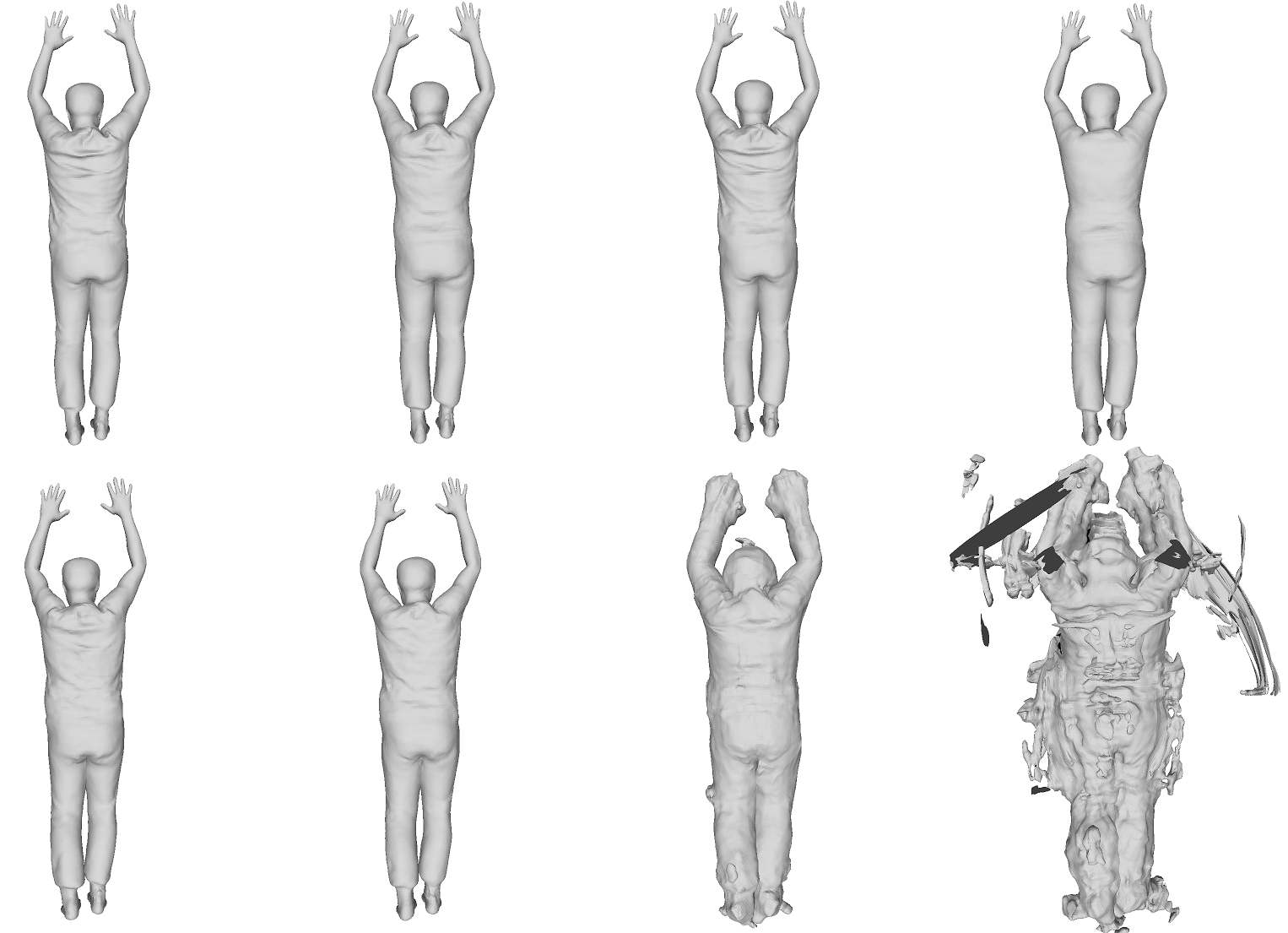}
\put(-230,-8){\small{100\% \qquad \qquad \; 50\% \qquad \qquad \quad 10\% \qquad \qquad \quad \; 5\% }}
\vspace{-5pt}
\caption{\small{
\textbf{Evaluation of pose encoding with different sizes of training data.} Top row: our local pose encoding. Bottom row: global pose encoding. While the global pose encoding suffers from severe overfitting artifacts, our local pose encoding generalizes well even if data size is severely limited.}}
\label{fig:eval_localpose}
\vspace{-10pt}
\end{figure}

Table \ref{tab:eavl_poseCondition} shows that our local pose conditioning achieves better reconstruction accuracy and fidelity for both interpolation and extrapolation. 
Note that the performance of global pose conditioning drastically degrades when the training data is reduced to less than 10\%, suffering from severe overfitting.
In contrast, our approach keeps roughly equivalent reconstruction accuracy even when only 5\% of the original training data is used, exhibiting few noticeable artifacts (see Fig.~\ref{fig:eval_localpose}).  

\begin{figure}[t]
\centering
\includegraphics[width=\linewidth]{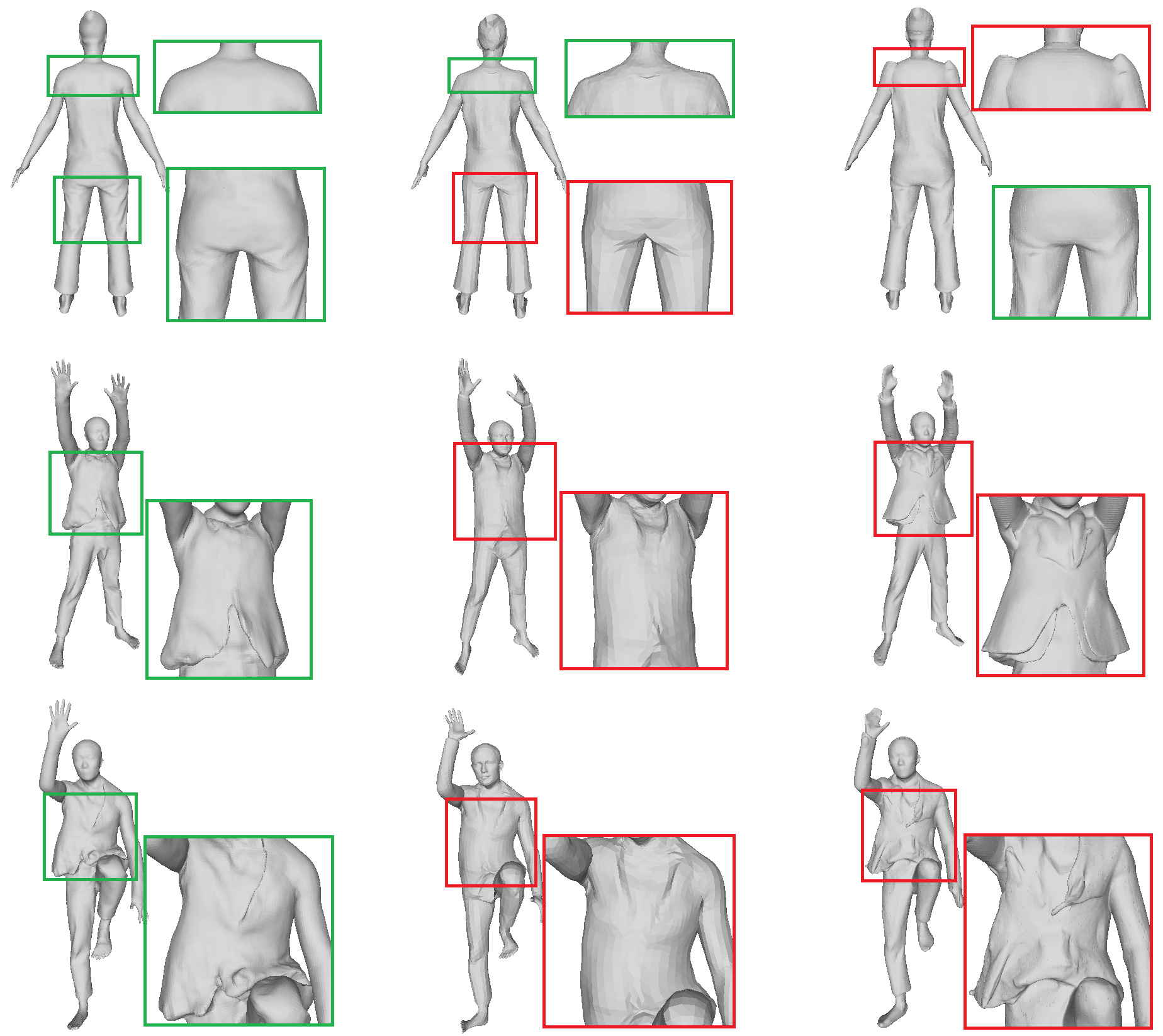}
\put(-210,-8){\small{Ours \qquad \qquad \qquad CAPE~\cite{CAPE:CVPR:20} \qquad \qquad \; NASA~\cite{deng2019neural}}}
\vspace{-5pt}
\caption{\small{
\textbf{Comparison with the SoTA methods.} We show qualitative results on the extrapolation task, illustrating the advantages of our method as well as the limitations of the existing approaches.}}
\label{fig:comp_sota}
\vspace{-10pt}
\end{figure}

\noindent
{\bf   Comparison with SoTA.}
We compare the proposed method with two state-of-the-art methods that also learn an articulated parametric human model with pose correctives from real world scans~\cite{deng2019neural,CAPE:CVPR:20}. CAPE~\cite{CAPE:CVPR:20} learns pose-dependent deformations on a fixed mesh topology using graph convolutions \cite{ranjan2018generatingcoma}, but requires surface registration for training. NASA~\cite{deng2019neural}, on the other hand, can be learned without registration but needs to determine occupancy values. We train both methods using registered CAPE data. Table \ref{tab:eval_comparison} shows that our approach achieves superior reconstruction accuracy and perceptual realism, while Fig.~\ref{fig:comp_sota} illustrates limitations of the prior methods. 
As CAPE relies on a template mesh with a fixed topology, the reconstructions are not only less detailed but also fail to capture topological changes such as the lifting up of the jacket.
While NASA can model pose-dependent shapes using articulated implicit functions, discontinuites and ghosting artifacts are visible, as the implicit functions of each body part are learned independently, which limits generalization to unseen poses. In contrast, our approach can produce highly detailed and globally coherent pose-dependent deformations without template-registration.

% \begin{table}[h]
%     \centering
%     \begin{tabular}{|c|p{1.5cm}|c|c|c|}
%     \hline
%         \multicolumn{2}{|c|}{}& Ours & CAPE & NASA\\
%         \hline
%         \multirow{4}{*}{Int.} & $D_{s2m}$ $\downarrow$ & \textbf{5.70} & 9.7 & 11.2 \\
%         \cline{2-5}
%         & $D_n$ $\downarrow$ & \textbf{0.253} & 0.308 & 0.289 \\
%         \cline{2-5}
%         & $P_i$ $\uparrow$ & \textbf{1.00} & 0.366 & 0.760\\
%         \cline{2-5}
%         & $P_v$ $\uparrow$ & \textbf{1.00} & 0.836 & 0.842\\
%         \hline
%         \multirow{2}{*}{Ex.} & $P_i$ $\uparrow$ & \textbf{1.00} & 0.273 & 0.522\\
%         \cline{2-5}
%         & $P_v$ $\uparrow$ & \textbf{1.00} & 0.730 & 0.653\\
%         \hline
%     \end{tabular}
%     \caption{Comparison of our method with SoTA pose-dependent shape models.}
%     \label{tab:eval_comparison}
% \end{table}

\begin{table}[t]
    \centering
    \caption{Comparison with the state-of-the-art pose-aware shape modeling methods.}
    \vspace{-0.1in}{\small
    \begin{tabular}{lp{1.5cm}ccc}
    \toprule
        \multicolumn{2}{c}{}& Ours & CAPE~\cite{CAPE:CVPR:20} & NASA~\cite{deng2019neural}\\
        \hline
        \multirow{4}{*}{Int.} & $D_{s2m}$ $\downarrow$ & \textbf{0.570} & 0.970 & 1.12 \\
        % \cline{2-5}
        & $D_n$ $\downarrow$ & \textbf{0.253} & 0.308 & 0.289 \\
        % \cline{2-5}
        & $P_i$ $\uparrow$ & \textbf{0.5} & 0.268 & 0.432\\
        % \cline{2-5}
        & $P_v$ $\uparrow$ & \textbf{0.5} & 0.455 & 0.457\\
        \hline
        \multirow{2}{*}{Ex.} & $P_i$ $\uparrow$ & \textbf{0.5} & 0.214 & 0.343\\
        % \cline{2-5}
        & $P_v$ $\uparrow$ & \textbf{0.5} & 0.422 & 0.395\\
        \bottomrule
    \end{tabular}}
    \label{tab:eval_comparison}
   % \vspace{-5pt}
\end{table}

\begin{figure}[h]
\centering
\includegraphics[width=\linewidth]{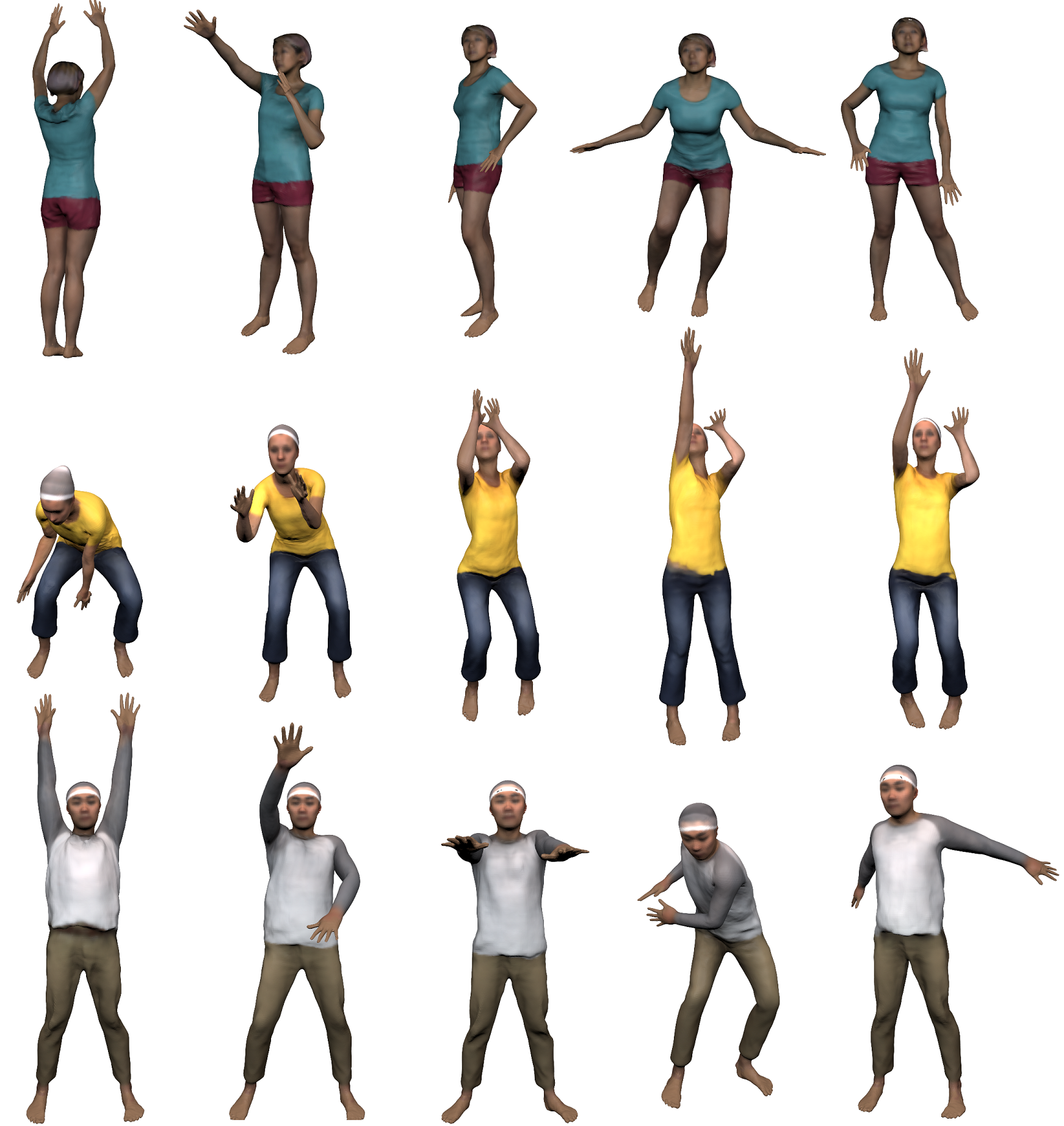}
\vspace{-20pt}
\caption{\small{
\textbf{Textured Scanimats.} Our method can be extended to texture modeling, allowing us to automatically build a Scanimat with high-resolution realistic texture.}}
% \textbf{Results with texture module.} }}
\label{fig:texture}
\vspace{-10pt}
\end{figure}

\noindent
{\bf   Learning a Fully Textured Avatar.}
We extend our pose-aware shape modeling to appearance modeling by predicting texture fields \cite{oechsle2019texture, saito2019pifu}; see Appendix~\ref{sup:sec:implementation_details} for details. Figure \ref{fig:texture} shows that high-resolution texture can be modeled without 2D texture mapping, which illustrates another advantage of eliminating the template-mesh requirement.

\section{Discussion and Future Work}
\label{sec:conclusion}

We introduced SCANimate, a fully automatic framework to create high-quality avatars (Scanimats), with realistic clothing deformations, driven by pose parameters, that are directly learned from raw 3D scans. Our experiments show that decomposing articulated deformations from scanned data is now possible in a weakly supervised manner by combining body-guided supervision with cycle-consistency regularization. Previously, the difficulty of accurate and coherent surface registration limited the field from analysing and modeling complex clothing deformations involving multiple garments from real-world observations. Our approach enables, for the first time, learning of physically plausible clothing deformations from raw scans, unlocking the possibility of realistic avatar learning from data. 

\noindent
{\bf   Limitations and Future Work.}
The current representation works well for clothing that is topologically similar to the body. 
The method may fail for clothing, like skirts, that deviates significantly from the body; see Appendix~\ref{sup:sec:discussion} for an example. 
Clothing wrinkles tend to be stochastic; that is, for a specific pose, they may differ depending on the preceding sequence of poses.
The current model, however, is deterministic.
Future work should factor the surface texture into albedo, shape, and lighting enabling more realistic relighting of Scanimats. 
Additionally, an adversarial texture loss \cite{huang2020advtex} could improve visual quality.
Here we model a person in a single garment. Learning a generative model with clothing variety should be possible but will require training data of varied clothing in varied poses.
Most exciting is the idea of fitting Scanimats to, or even learning them from, images or videos.
Finally, extending this approach to model hand articulation and facial expressions should be possible using expressive body models like SMPL-X \cite{pavlakos2019expressive}.

\revision{
\noindent
{\footnotesize 
{\bf   Acknowledgements}: Q.~Ma was partially funded by the Deutsche Forschungsgemeinschaft (DFG, German Research Foundation) - 276693517 SFB 1233.
% \noindent
{\bf   Disclosure:} MJB has received research gift funds from Adobe, Intel, Nvidia, Facebook, and Amazon. While MJB is a part-time employee of Amazon, his research was performed solely at, and funded solely by, Max Planck. MJB has financial interests in Amazon, Datagen Technologies, and Meshcapade GmbH.
}
}
\clearpage 

{\small
	\bibliographystyle{ieee_fullname}
	\balance
	\bibliography{paper}
}
\clearpage

%%%%%%%% Supplementary
\appendix
{\noindent\Large\textbf{Appendix}}
\setcounter{page}{1}
\counterwithin{figure}{section}
\counterwithin{table}{section}

\section{Implementation Details}\label{sup:sec:implementation_details}
\subsection{Network Architectures}
Our forward and inverse skinning networks are based on multi-layer perceptrons, where the intermediate neuron size is $(256, 256, 256, 24)$ with a skip connection from the input feature to the $2$nd layer, and nonlinear activations using LeakyReLU except the last layer that uses softmax to obtain normalized skinning weights. As an input, we take the Cartesian coordinates of a queried location, which is encoded into a high dimensional feature using the positional encoding \cite{mildenhall2020nerf} with up to $6$-th and $8$-th order Fourier features for the forward skinning net $g^c_{\Theta_1}(\cdot)$ and the inverse skinning net $g^s_{\Theta_2}(\cdot)$, respectively. Note that the inverse skinning net $g^s_{\Theta_2}(\cdot)$ takes a latent embedding $\boldsymbol{z}^s_i \in \mathbb{R}^{64}$ as an additional input in order to learn the skinning weights of scans in different poses. 

To model the geometry of clothed humans in a canonical pose, we also use a multi-layer perceptron $f_\Phi(\cdot)$, where the intermediate neuron size is $(512, 512, 512, 343, 512, 512, 1)$ with a skip connection from the input feature to the $4$th layer, and nonlinear activations using softplus with $\beta=100$ except the last layer as in \cite{gropp2020implicit}. The input feature consists of the Cartesian coordinates of a queried location, which are encoded using the positional encoding of up to $8$-th order Fourier features, and the localized pose encoding in $\mathbb{R}^{92}$. The texture inference network uses the same architecture as the geometry module $f_\Phi(\cdot)$ except the last layer with $3$ dimensional neurons for color prediction, and the input layer replaced with the concatenation of the same input and the second last layer of $f_\Phi(\cdot)$ so that the color module is aware of the underlying geometry. 
 
\subsection{Training Procedure}
Our training consists of three stages. First, we pretrain $g^c_{\Theta_1}(\cdot)$ and $g^s_{\Theta_2}(\cdot)$ with the following relative weights:
$\lambda_{B}=10.0$,
$\lambda_{S}=1.0$,
$\lambda_{C'}=0.0$,
$\lambda_{C''}=0.0$,
$\lambda_{Sp}=0.001$,
$\lambda_{Sm}=0.0$,
and $\lambda_{Z}=0.01$.
After pretraining, we jointly train $g^c_{\Theta_1}(\cdot)$ and $g^s_{\Theta_2}(\cdot)$ using the proposed cycle consistency constraint with the following weights:
$\lambda_{B}=10.0$,
$\lambda_{S}=1.0$,
$\lambda_{C'}=1.0$,
$\lambda_{C''}=1.0$,
$\lambda_{Sp}=0.001$,
$\lambda_{Sm}=0.1$,
and $\lambda_{Z}=0.01$. We multiply $\lambda_{C''}$ by $10$ for the second half of the training iterations. For the two stages above, we use $6890$ points of the SMPL vertices and $8000$ points uniformly sampled on the scan data, which is dynamically updated at every iteration. 

Once the training of the skinning networks is complete, we fix the weights of $g^c_{\Theta_1}(\cdot)$, $g^s_{\Theta_2}(\cdot)$, and $\left\{ \boldsymbol{z}^s_i \right\}$, and train the geometry module $f_\Phi(\cdot)$ with the following hyper parameters:
$\lambda_{igr}=1.0$,
$\lambda_{o}=0.1$,
and $\alpha=100$. To compute $E_{LS}$, we uniformly sample $5000$ points on the scan surface at each iteration. We compute $E_{IGR}$ by combining $2000$ points within a bounding box and $10000$ points perturbed with the standard deviation of $10$cm from the surface geometry, half of which is sampled from the scans and the remaining from the SMPL body vertices. Note that $E_O$ uses only $2000$ points sampled from the bounding box to avoid overly penalizing zero crossing near the surface. 

We train each stage with the Adam optimizer with learning rates of $0.004$, $0.001$, and $0.001$, respectively. They are decayed by the factor of $0.1$ at $1/2$ and $3/4$ of the training iterations.
The first stage runs for 80 epochs and the second for 200 epochs.

\subsection{Texture Inference}
To model texture on the implicit surface, we model texture fields parameterized by a neural network, denoted as $f^c(\boldsymbol{x}): \mathbb{R}^3 \to \mathbb{R}^3$, following \cite{oechsle2019texture, saito2019pifu}. Given the ground-truth color $c(\boldsymbol{x})$ at a location $x$ on the surface, we learn the network weights of $f^c(\cdot)$ by minimizing the $L1$ reconstruction loss: $\left|f^c(\boldsymbol{x}) - c(\boldsymbol{x})\right|$. We sample $5000$ points from the input scans at every iteration and optimize using the Adam optimizer with a learning rate of $0.001$ and the same decay schedule as the geometry module. We train the texture module for $1.8 M$ iterations.

\subsection{Other Details}
\paragraph{Concave Region Detection.} 
We exclude concave regions from the smoothness constraint to avoid propagating incorrect skinning weights at the self-intersection regions. We detect them by computing the mean curvature on the surface of scans with the threshold of $0.2$. Note that while we empirically find our detection algorithm is sufficient for our training data, utilizing external information such as body part labels is possible when available to improve robustness.

\paragraph{Obtaining Canonical Body.}
The canonicalized body $\mathbf{B}^c_i$ in Eq.~\ref{eq:body_guided} is a body model of the subject in a canonical pose with pose dependent deformations. We obtain the pose correctives by activating pose-aware blend shapes in the SMPL model~\cite{loper2015smpl} given the body pose $\theta$ at frame $i$.

\paragraph{Removing Distorted Triangles.}
When the input scans are canonicalized, triangle edges that belong to self-intersection regions are highly distorted. As these regions must be separated in the canonical pose, we remove all triangles for which any edge length is larger than its initial length multiplied by $4$.

\section{Discussions}\label{sup:sec:discussion}

We provide additional discussion to clarify technical details in the main text, including choice of latent autodecoding and the similarity of forward and inverse skinning networks, and discussion over failure cases.

\subsection{Latent Autodecoding}

The purpose of learning $g^s(\cdot,\boldsymbol{z})$ is to stably canonicalize raw scans. To this end, we use auto-decoding $\boldsymbol{z}$ as in \cite{park2019deepsdf} for the following advantages. 
Auto-decoding self-discovers the latent embedding $\boldsymbol{z}$ such that the loss function is minimized, allowing the network to better distinguish each scan regardless of the similarity in the pose parameters. Thus, $\boldsymbol{z}$ can implicitly encode not only pose information but also anything necessary to distinguish each frame. Furthermore, due to no dependency on pose parameters, auto-decoding is more robust to the fitting error of the underlying body model. As a baseline we replace autodecoding by regressing skinning weights on pose parameters of a fitted SMPL body. We use the energy function $E_{cano}$ in Eq.~\ref{eq:cano} without the term of $E_Z$ to evaluate the performance of the two. While pose regression results in 0.043, autodecoding achieves a much lower local minimum at 0.025, showing superior performance against the baseline.

\subsection{CAPE Dataset Limitation}
\begin{wrapfigure}{R}{0.45\linewidth}
  \begin{center}
    \includegraphics[width=\linewidth]{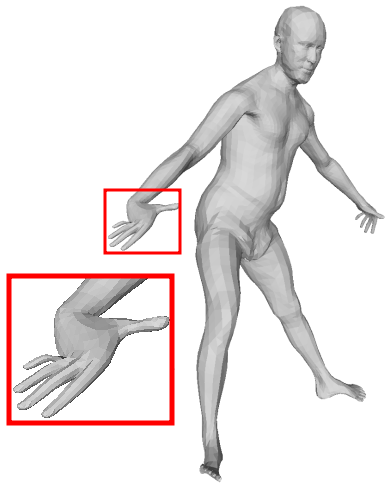}
  \end{center}
\end{wrapfigure}
Some frames of the CAPE dataset~\cite{CAPE:CVPR:20} contain erroneous body fitting around the wrists and ankles, as shown in the right inset figure, resulting in unnecessary distortions around the regions. 
Due to the smoothness regularization in our method, such a distortion can be propagated to the nearby regions, and hence a larger region may be discarded. However, the proposed shape learning method complements such a missing region from other canonicalized scans, and our reconstructed Scanimats do not suffer from the small errors in pose fitting.

\subsection{Combining Skinning Networks}

As in Eq.~\ref{eq:skin_net}, $g^c$ and $g^s$ are formulated separately. 
This is in accordance with the idea of predicting skinning weights for both forward and backward transformations.
However, if one considers the skinning networks in another point of view, particularly when regarding them as mappings from 3D space coordinates conditioned on different frames to skinning weights, it is clear that $g^c$ is a special case of $g^s$.
Thus in practical implementation, one can either set up two networks corresponding to $g^c$ and $g^s$, respectively, or set up a single networks in an autodecoder manner with a single common latent vector $\boldsymbol{z}^c$ for all the forward skinning weights prediction and per-frame latent vectors $\boldsymbol{z}^s_i$ for inverse skinning weights prediction in each posed frame.

\subsection{Failure Cases}
\begin{figure}
    \centering
    \includegraphics[width=\linewidth]{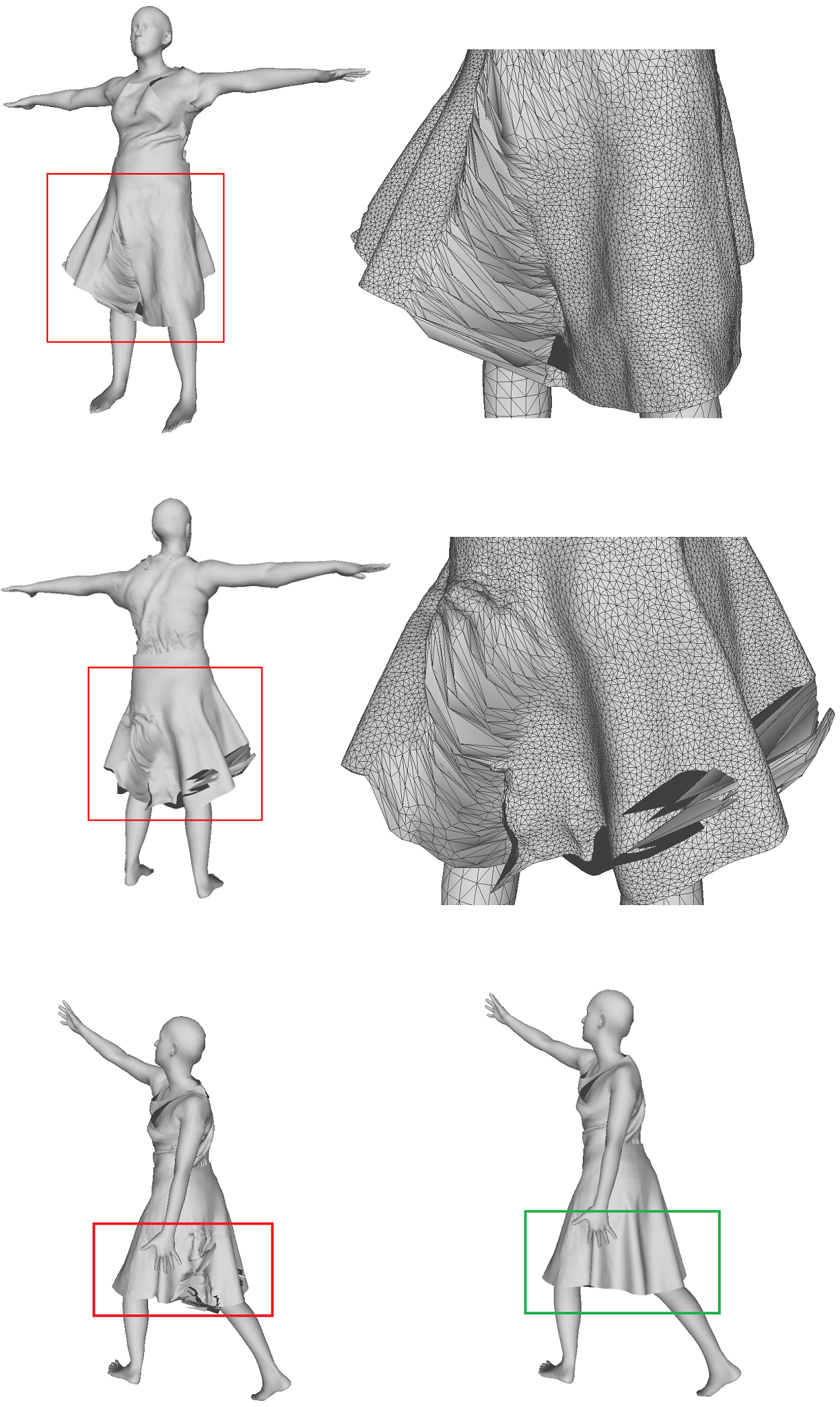}
    \put(-230,267){\small{Canonicalized scan (frontal) \qquad \; Zoomed-in view on the skirt}}
    \put(-230,132){\small{Canonicalized scan (back) \qquad \quad Zoomed-in view on the skirt}}
    \put(-220,-8){\small{Predicted reposed scan \qquad \quad \; Ground truth posed scan}}
    \caption{Failure cases of canonicalizing a clothed human with a synthetic skirt. We show surface triangles in the zoomed-in images to highlight the severe stretching artifacts of the skirt between legs.}
    \vspace{-10pt}
    \label{fig:failure_case}
\end{figure}

As mentioned in the main paper, while the current pipeline performs well for clothing that is topologically similar to the body, the method may fail for clothing, like skirts, whose topology may deviate significantly. Fig. ~\ref{fig:failure_case} shows a failure case of canonicalizing a person with a skirt synthetically generated using a physics-based simulation. The SMPL-guided initialization of skinning weights fails recovering from poor local minima. We leave for future work a garment-specific tuning of hyperparameters and more robust training schemes for various clothing types.

% \clearpage

\section{Additional Qualitative Results}
\begin{figure}[htb]
\centering
\includegraphics[width=0.9\linewidth]{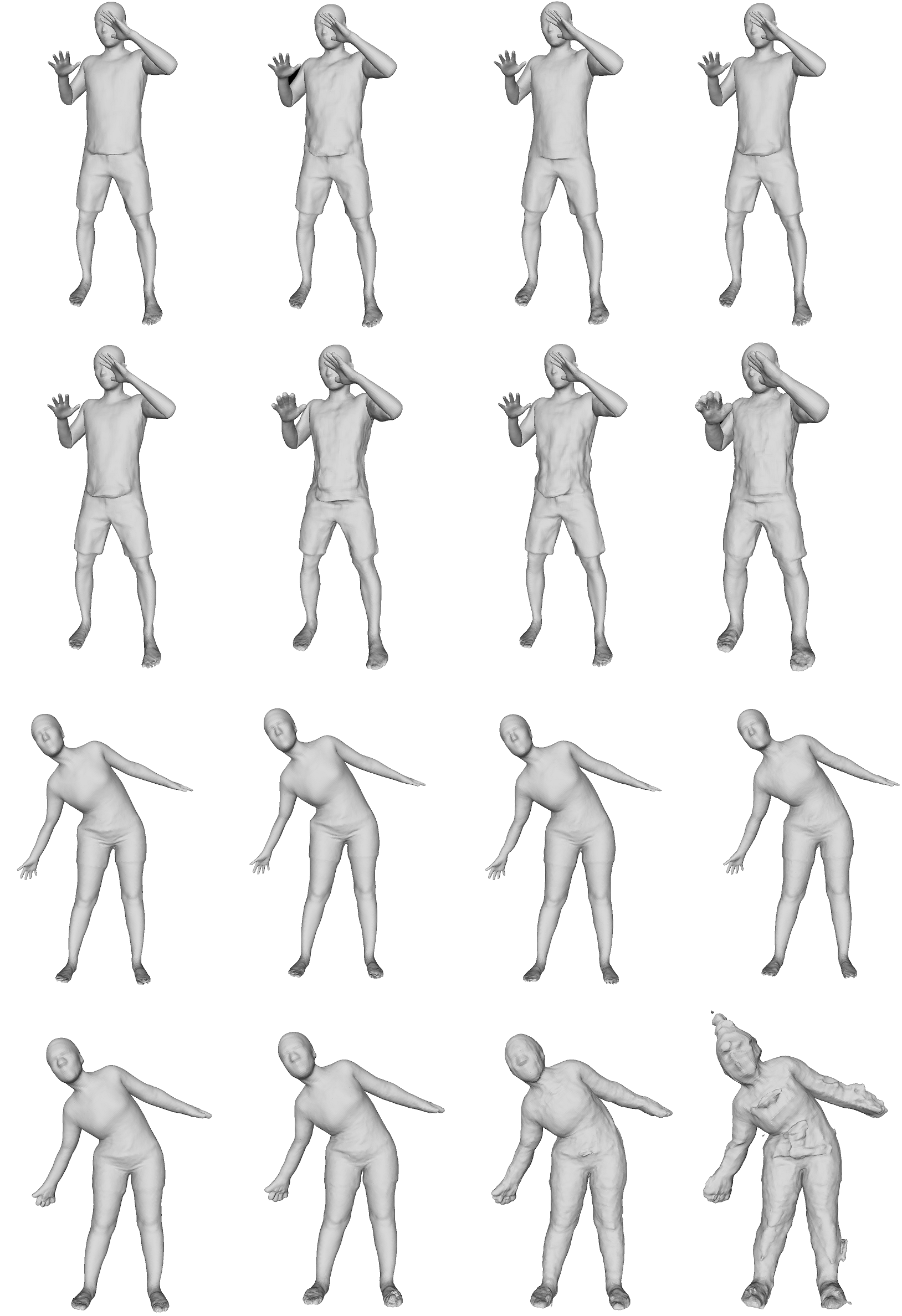}
\put(-198,-8){\small{100\% \qquad \qquad \; 50\% \qquad \quad \; 10\% \qquad \qquad 5\% }}
\vspace{-5pt}
\caption{\small{
\textbf{Evaluation of pose encoding with different sizes of training data.} Top row: our local pose encoding. Bottom row: global pose encoding. While the global pose encoding suffers from severe overfitting artifacts, our local pose encoding generalizes well even if data size is severely limited.}}
\label{fig:eval_localpose_sup}
\vspace{-10pt}
\end{figure}

% Please watch the video at {\small\url{https://scanimate.is.tue.mpg.de}} for animated results.
\paragraph{Locally Pose-aware Shape Learning.} Fig.~\ref{fig:eval_localpose_sup}, an extended figure of Fig.~\ref{fig:eval_localpose} in the main paper, shows more qualitative comparison on pose encoding with different sizes of training data.

\begin{figure}[htb]
\centering
\includegraphics[width=\linewidth]{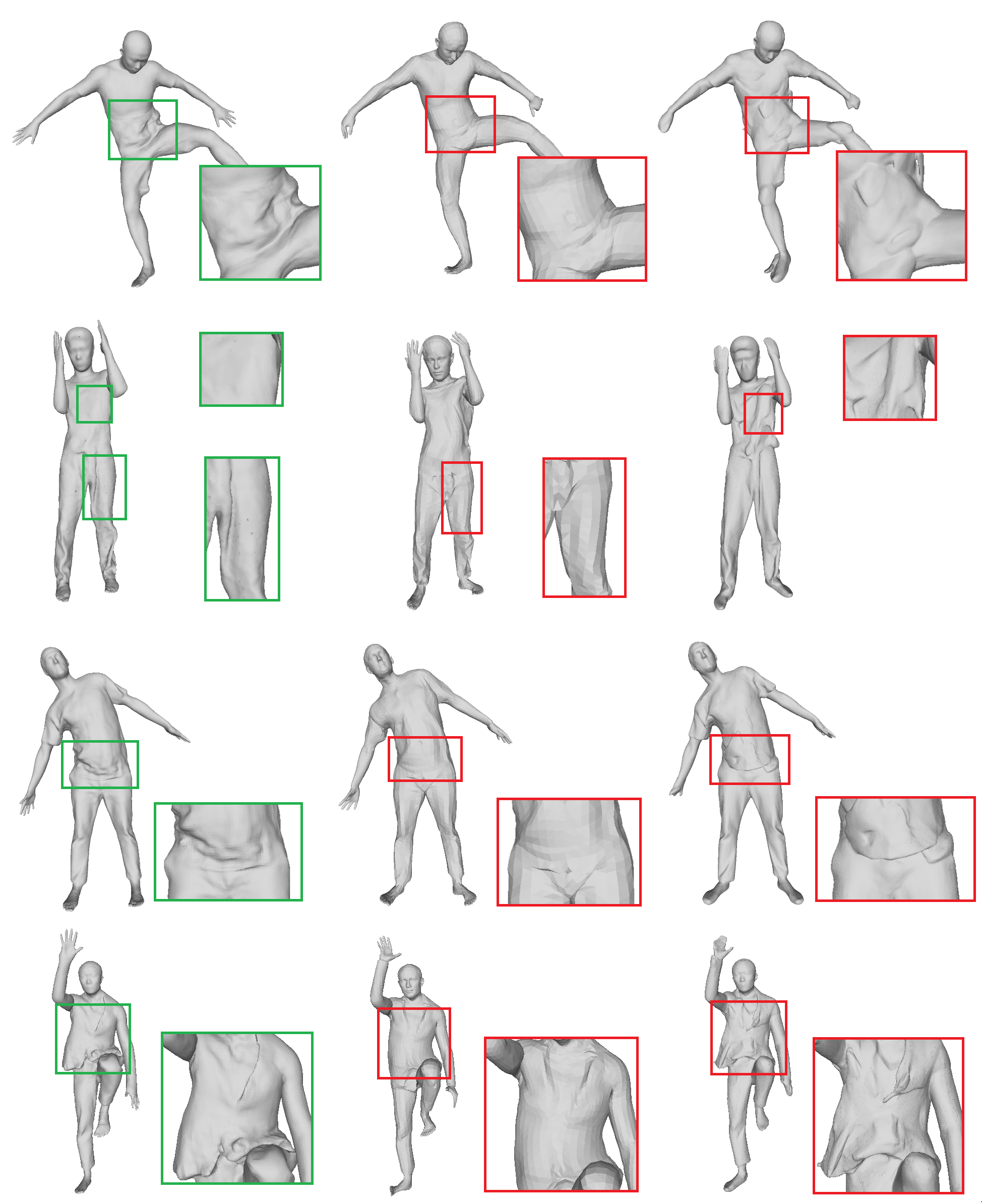}
\put(-210,-8){\small{Ours \qquad \qquad \qquad CAPE~\cite{CAPE:CVPR:20} \qquad \qquad \; NASA~\cite{deng2019neural}}}
\vspace{-5pt}
\caption{\small{
\textbf{Comparison with the SoTA methods.} We show qualitative results on the extrapolation task, illustrating the advantages of our method as well as the limitations of the existing approaches.}}
\label{fig:comp_sota_sup}
\vspace{-10pt}
\end{figure}

\paragraph{Comparison with the SoTA methods.} Fig.~\ref{fig:comp_sota_sup}, an extended figure of the main paper Fig.~\ref{fig:comp_sota}, shows more qualitative comparison with the SoTA methods.

\begin{figure}[htb]
\centering
\includegraphics[width=\linewidth]{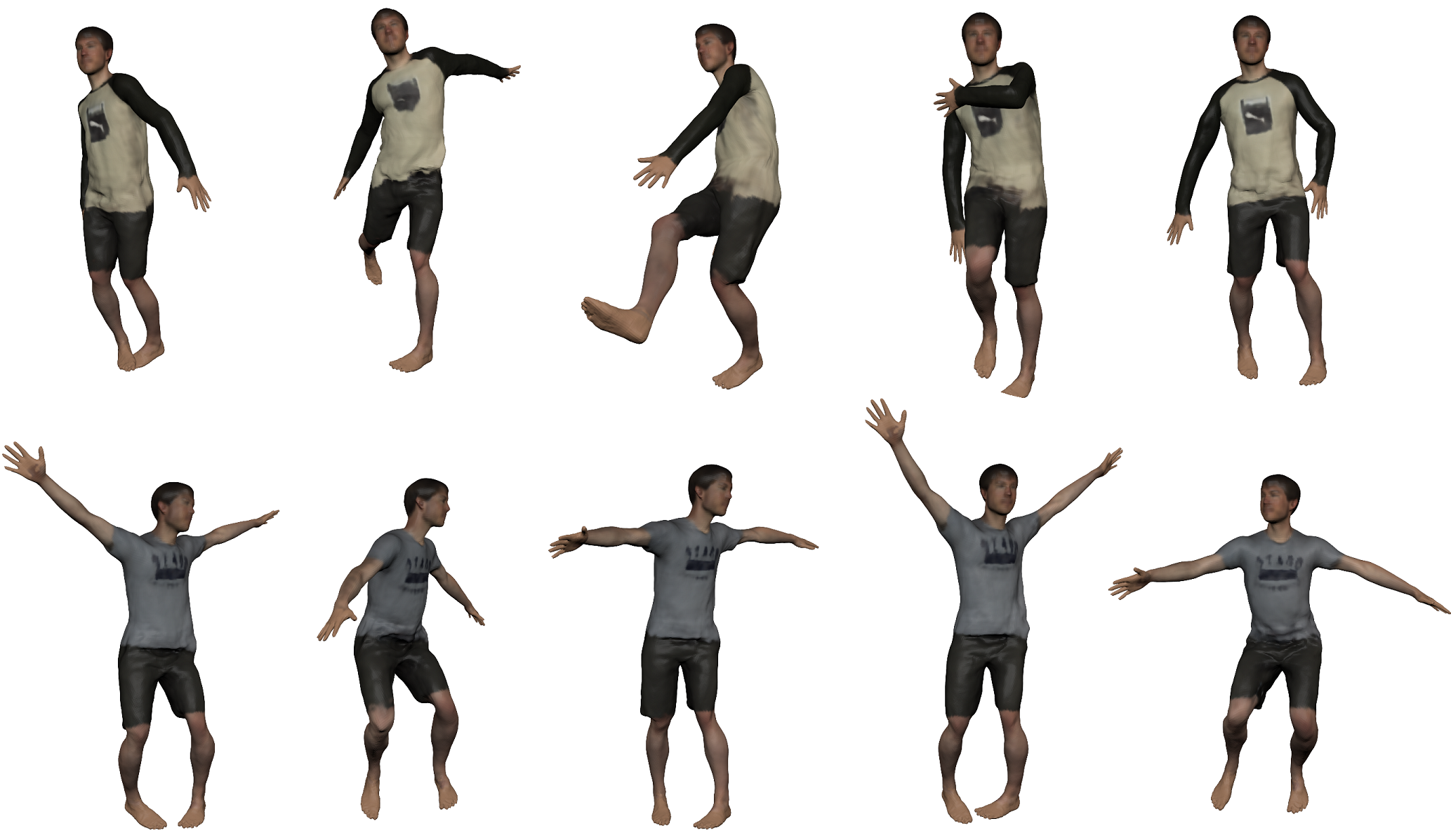}
\vspace{-10pt}
\caption{\small{
\textbf{Textured Scanimats.} Our method can be extended to texture modeling, allowing us to automatically build a Scanimat with high-resolution realistic texture.}}
\label{fig:texture_sup}
\vspace{-10pt}
\end{figure}

\paragraph{Textured Scanimats.} Fig.~\ref{fig:texture_sup}, an extended figure of Fig.~\ref{fig:teaser}, shows more examples of textured Scanimats.

\end{document}